\documentclass[journal]{IEEEtran}

\ifCLASSINFOpdf
   \usepackage[pdftex]{graphicx}
   \graphicspath{{../pdf/}{../jpeg/}}
   \DeclareGraphicsExtensions{.pdf,.jpeg,.png}
\else
   \usepackage[dvips]{graphicx}
   \graphicspath{{../eps/}}
   \DeclareGraphicsExtensions{.eps}
\fi

\date{January 11th 2022}

\usepackage[caption=false,font=footnotesize]{subfig}
\usepackage{stfloats}
\usepackage{xcolor}

\begin{document}

\title{Artificial Intelligence Generated Coins for Size Comparison}

\author{Gerald~Artner}



\maketitle

\begin{abstract}%
Authors of scientific articles use coins in photographs as a size reference for objects.
For this purpose, coins are placed next to objects when taking the photo.
In this letter we propose a novel method that uses artificial intelligence (AI) generated images of coins to provide a size reference in photos.
The newest generation is able to quickly generate realistic high-quality images from textual descriptions.
With the proposed method no physical coin is required while taking photos.
Coins can be added to photos that contain none.
Furthermore, we show how the coin motif can be matched to the object.
\end{abstract}

\maketitle

\textcolor{red}{This is an English translation of the original German article.
The English version is archived on arxiv.org with permission.
Please cite the original German version as:}

\textcolor{red}{Gerald Artner, ``Mit künstlicher Intelligenz generierte Münzen für Größenvergleiche,'' Mitteilungen der Österreichischen Numismatischen Gesellschaft, vol. 62, no. 2, pp. 9-16, 2022.
}

\section{Introduction}\label{sec:Introduction}
\begin{figure}[!ht]
\centering
\subfloat[]{\includegraphics[width=0.24\textwidth]{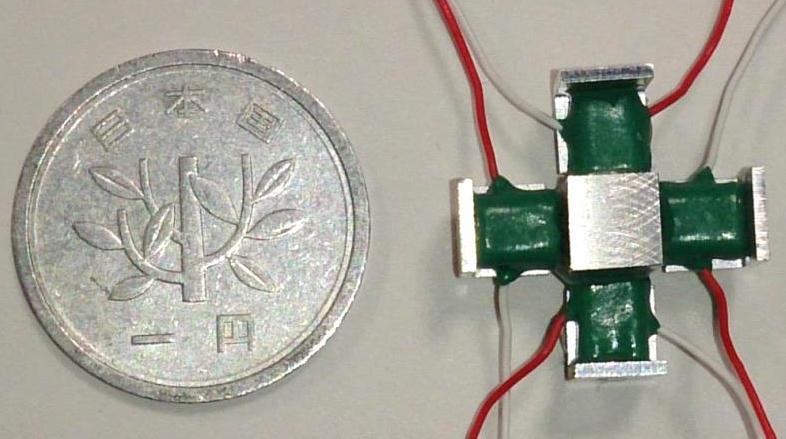}\label{fig_first_case}} ~
\subfloat[]{\includegraphics[width=0.24\textwidth]{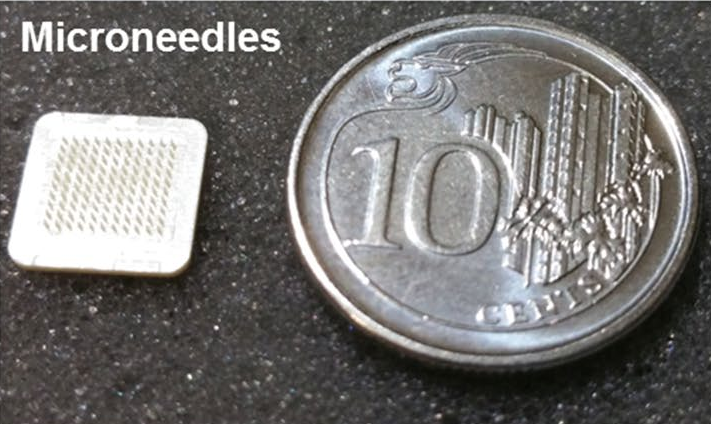}\label{fig_second_case}} \\
\subfloat[]{\includegraphics[width=0.24\textwidth]{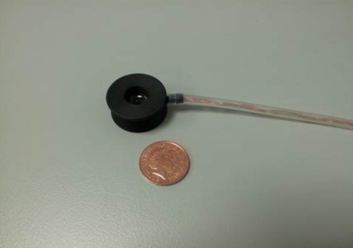}\label{fig_third_case}} ~
\subfloat[]{\includegraphics[width=0.24\textwidth]{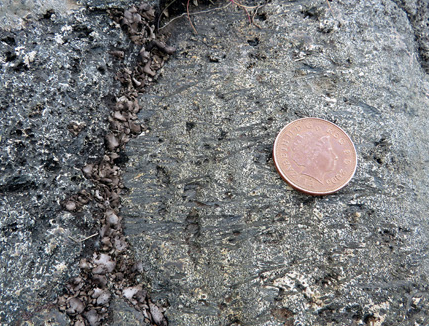}\label{fig_fourth_case}}
\caption{Typical examples where real coins are used as size reference in the scientific literature.
a) A 1 JPY coin compared to a positioner \cite{Muraoka2015}.
b) Microneedles with a coin that illustrates device scale \cite{Yeo2017}.
c) A coin provides an idea of the size of a medical elasticity probe \cite{Lee2016}.
d) A coin used for scale in geology \cite{Filippi2019}.
All images used with permission, Open Access under CC-BY \cite{CreativeCommons}.}
\label{fig_real_coins}
\end{figure}

Authors like to use objects of known size as references in photographs when the size of the model is not obvious to the viewer.
In engineering subjects, coins are most often used as size references for prototypes.
Typical examples with real coins are shown in Fig.~\ref{fig_real_coins}.
It can be argued that modern circulating coins have a standardized size and thus their use is not just a size reference but a measurement \cite{Artner2020}.
However, most authors use coins as a size reference or as a size comparison, for example \emph{``A quarter-dollar coin is presented for size comparison''} \cite{Zhang2020}, \emph{``Side and top views of our final benchmark boat (Benchy) print cured using monovoxel excitation printing, sitting on a dime for scale.''} \cite{Sanders2022} or \emph{`` [...] the designed filter and the waveguide are just bigger than a coin [...]''} \cite{Souza2017}.

Novel artificial intelligence (AI) imaging techniques have been applied in numismatics mainly to digitize, identify \cite{Capece2016,Kiourt2021,Zambanini2008,Kampel2008} and analyze \cite{Heinecke2021} coins.
Deep learning and artificial neuronal networks are also in use for grading and estimating the value of collectors coins \cite{Pan2018,Blance2021}.
Generative Adversarial Networks reconstruct images of damaged coins in \cite{Zachariou2020}.

\section{Proposed Method and Feasibility}

\begin{figure}[ht]
\centering
\subfloat[]{\includegraphics[width=0.24\textwidth]{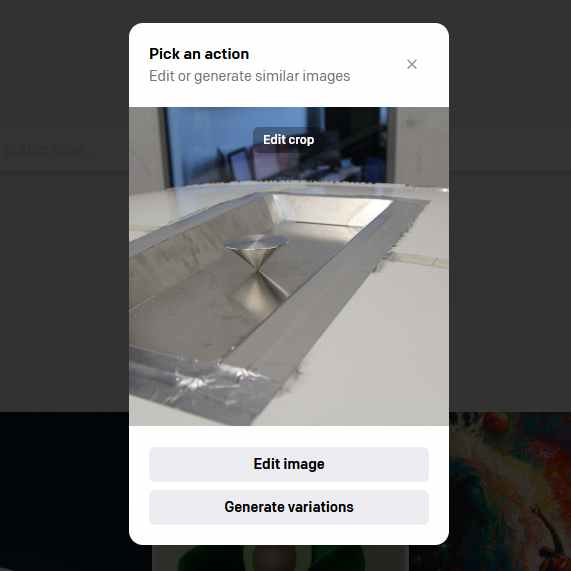}\label{fig_crop}} ~
\subfloat[]{\includegraphics[width=0.24\textwidth]{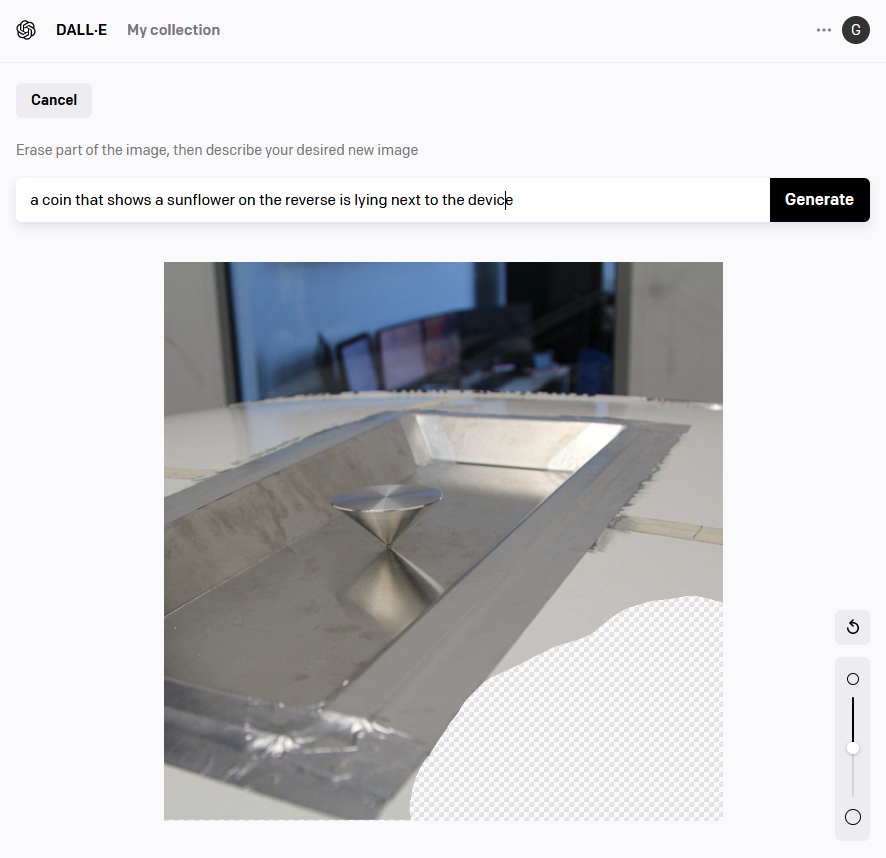}\label{fig_prompt}}
\caption{Tutorial for image editing with DALL-E 2
a) Uploaded images need to be cropped to squares. After cropping click on "Edit image". 
b) Use the box to provide a text prompt. Use the brush tool to clear the area where the coin will be placed. The image generator will inpaint the removed area.}
\label{fig_tutorial}
\end{figure}

\begin{figure}[ht]
\centering
\subfloat[]{\includegraphics[width=0.24\textwidth]{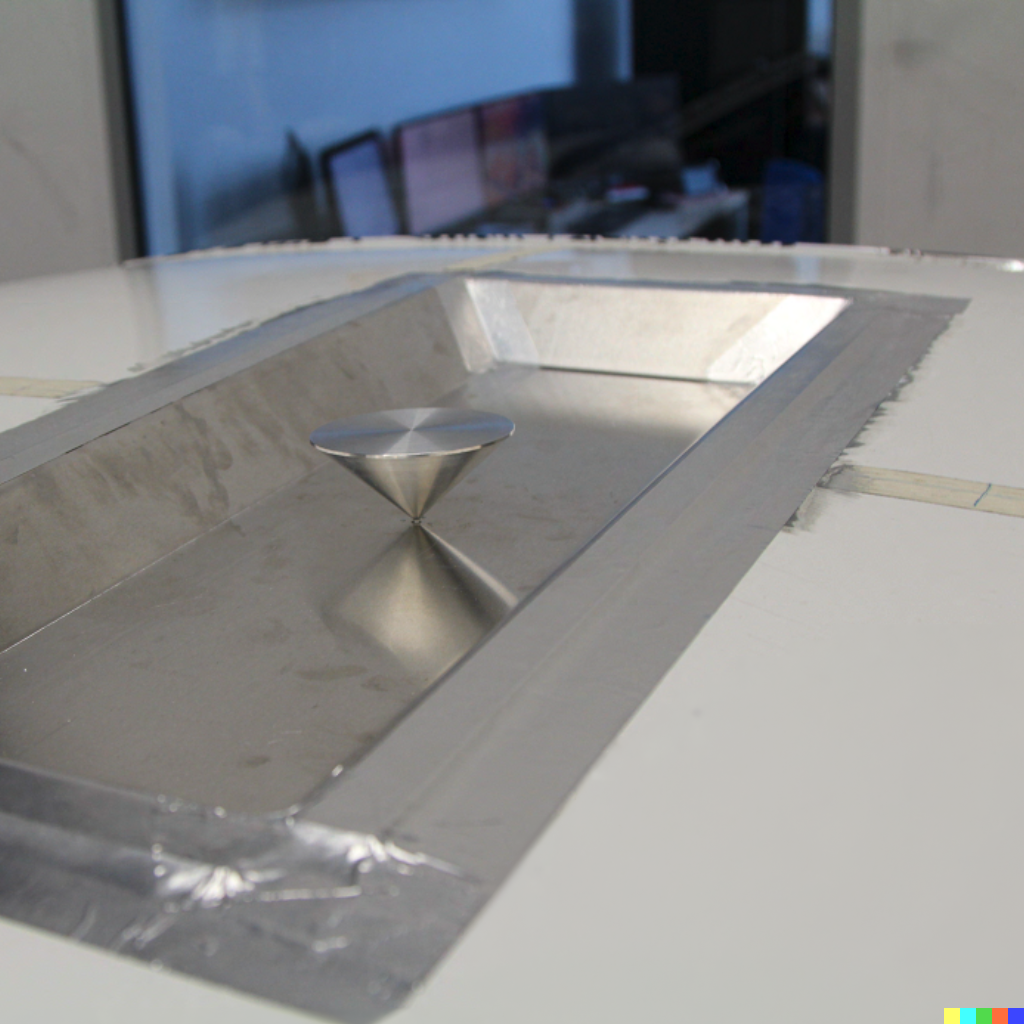}\label{fig_cavity_empty}} ~
\subfloat[]{\includegraphics[width=0.24\textwidth]{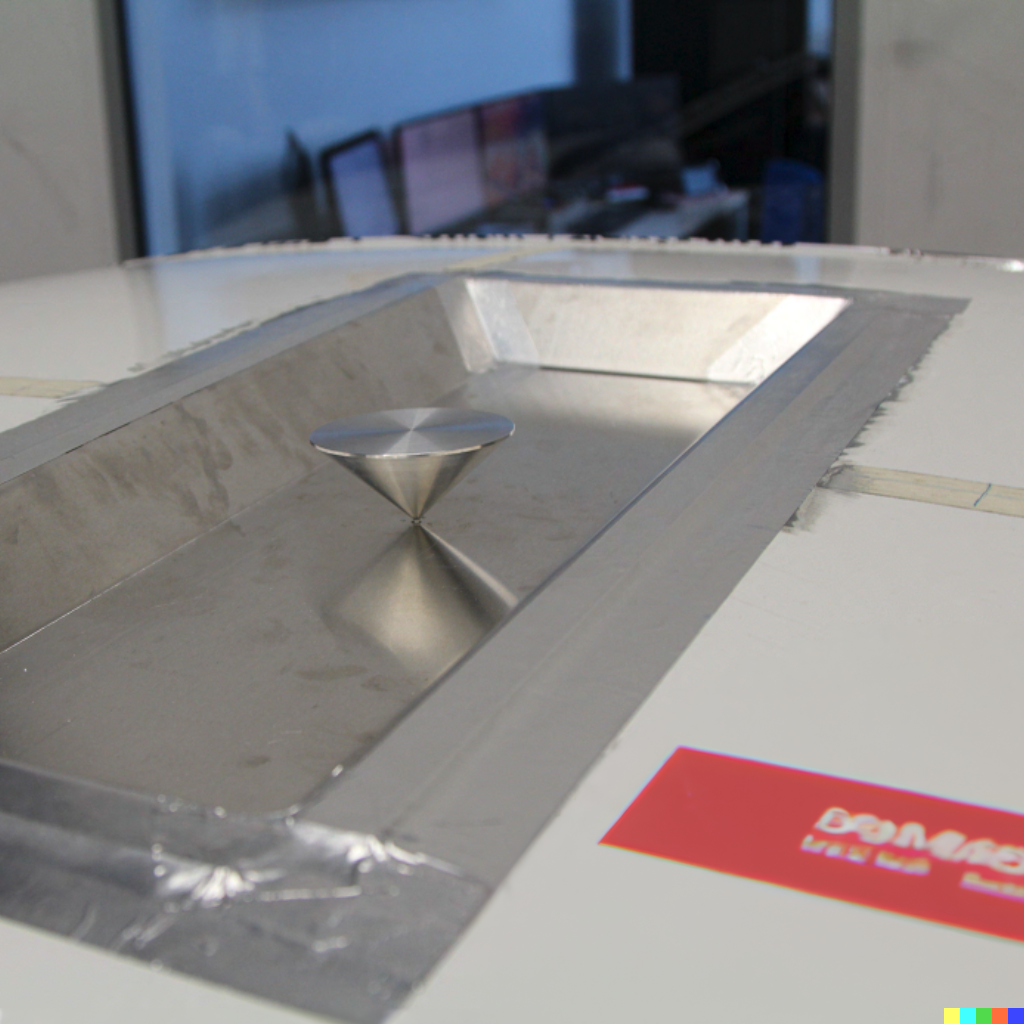}\label{fig_cavity_label}}
\caption{Sometimes undesirable images are generated.
a) Many outputs are inpainted space without a coin. 
b) Sometimes the system creates nonsensical text labels instead of coins 
All images were created with DALL-E 2 \cite{DALL-E}.
The original image was used with permission \cite{Artner2019IEEEAccess}, CC-BY.}
\label{fig_problems}
\end{figure}

\begin{figure}[ht]
\centering
\subfloat[]{\includegraphics[width=0.24\textwidth]{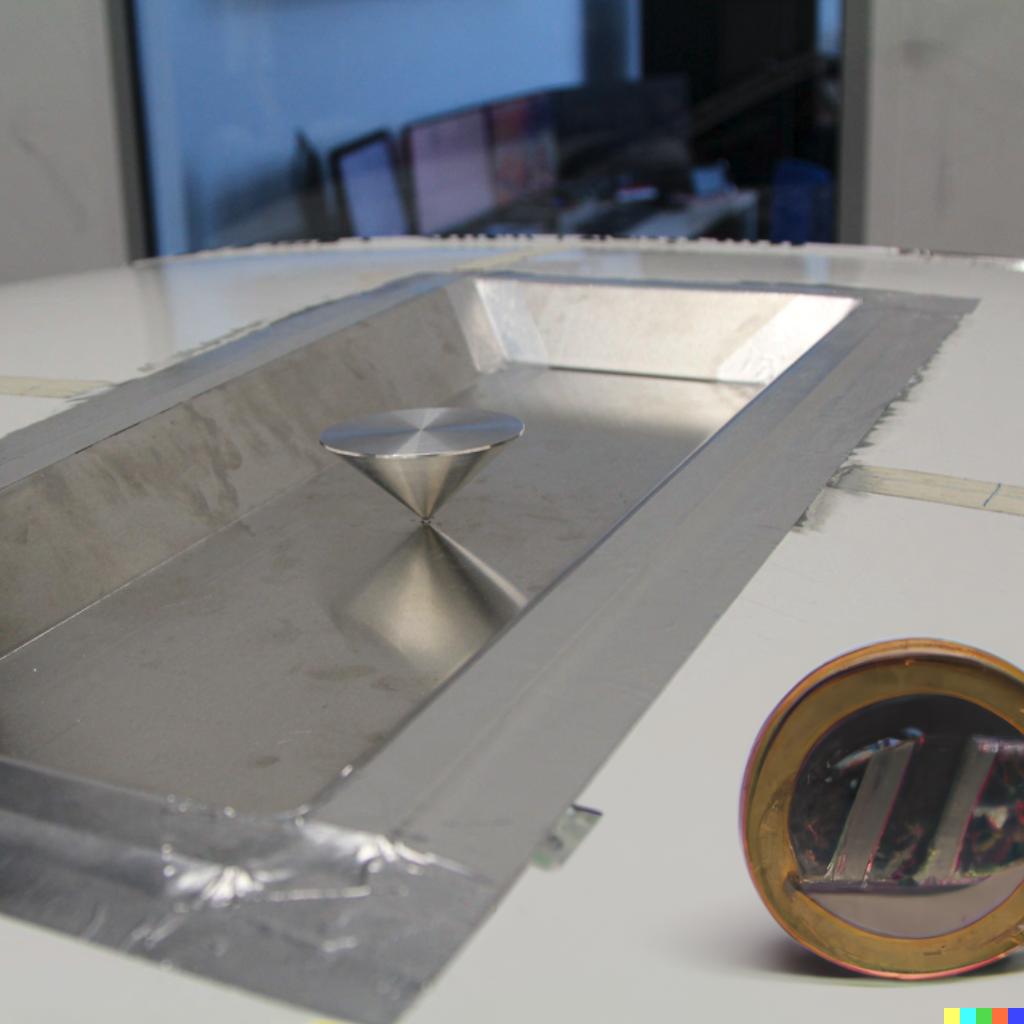}\label{fig_cavity_coin}} ~
\subfloat[]{\includegraphics[width=0.24\textwidth]{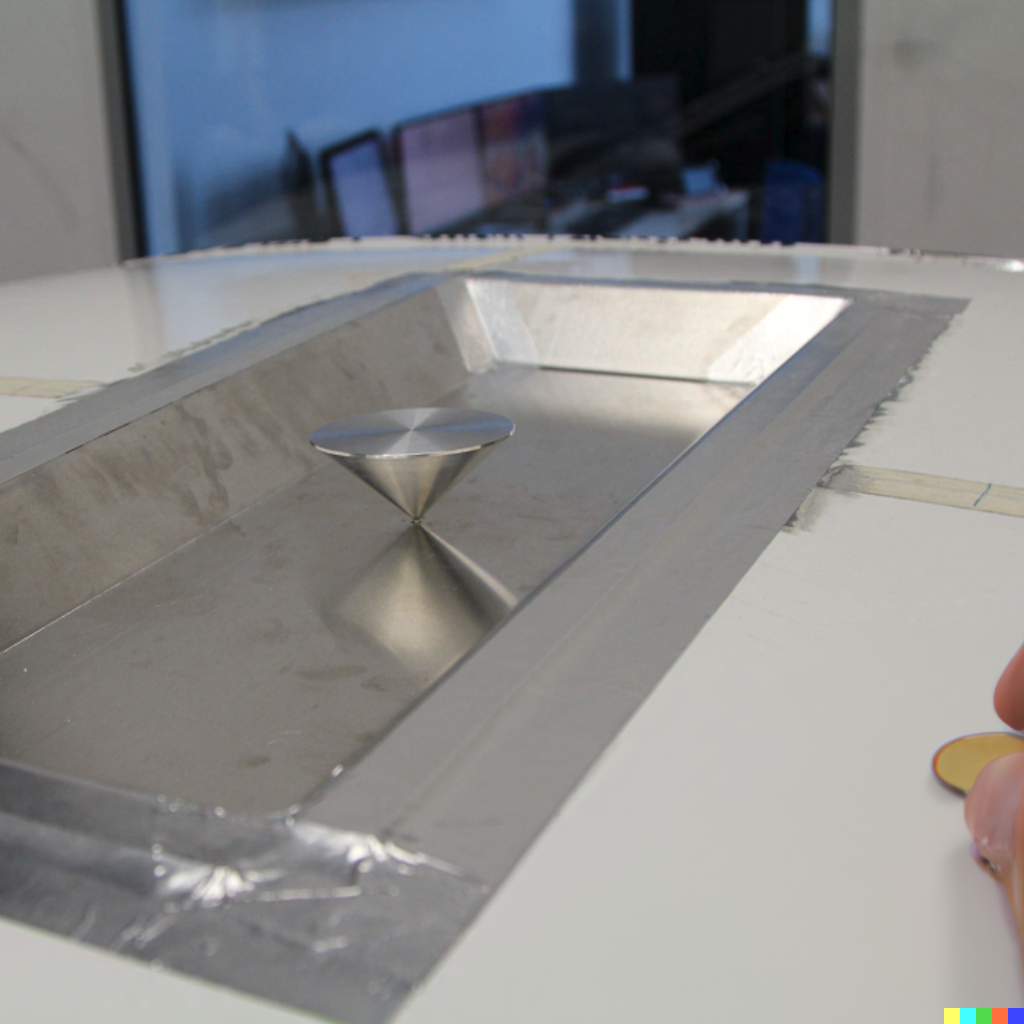}\label{fig_cavity_hand}}\\
\subfloat[]{\includegraphics[width=0.24\textwidth]{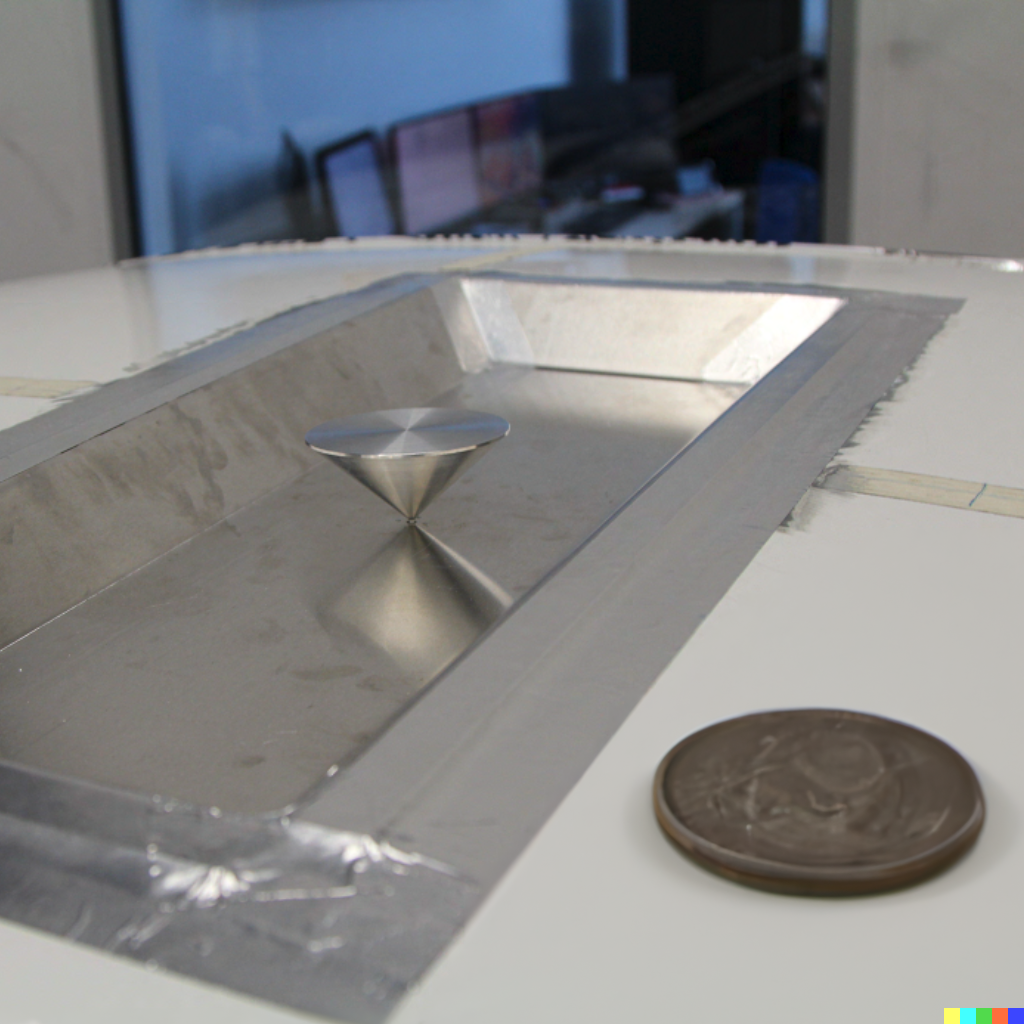}\label{fig_cavity_old_coin}} ~
\subfloat[]{\includegraphics[width=0.24\textwidth]{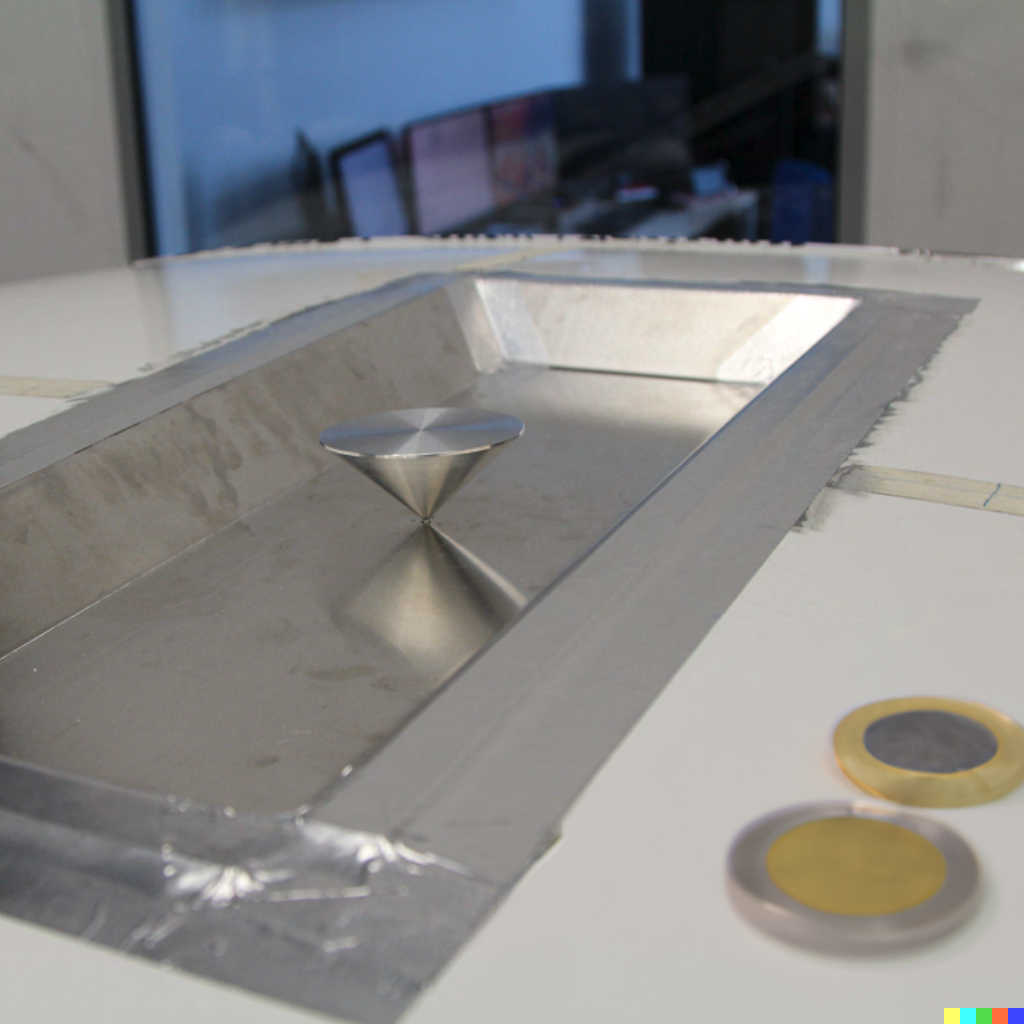}\label{fig_cavity_two_coins}}
\caption{AI generated images with coins that provide a size estimate. The bottom images show undesired results. 
a) A coin next to an antenna in a car roof cavity, prompt: ``a Euro coin lies next to the device, numismatics, reverse''.
b) The AI understood that the coin is actively placed there and generated an image of fingertips that place a coin, prompt: ``add a coin that shows a sunflower on the reverse''.
c) ``a historic coin is lying next to the device, numismatics, obverse''.
d) ``a Euro coin lies next to the device, numismatics, reverse''.
}
\label{fig_results}
\end{figure}

Recently developed text to image generators have become increasingly simple to use \cite{Frolov2021}.
They can be accessed via web interfaces and require no knowledge of machine learning algorithms.
We demonstrate how they can be used to add synthetic images of coins for size comparison.
We use the hierarchical text-conditional image generation system DALL-E 2 that has recently been opened to a wider public \cite{DALL-E}.
The OpenAI system DALL-E 2 uses a diffusion based method with the  Contrastive Language-Image Pre-training (CLIP) model \cite{Ramesh2020,Ramesh2022,Radford2021}.
Its language model is based on the Generative Pre-trained Transformer 3 (GPT-3) \cite{Brown2020}.
Fig.~\ref{fig_tutorial} shows screenshots of the cropping and drawing/promt steps.
To add a coin to a photo:
\begin{enumerate}
\item Upload the desired image on the DALL-E system. The image needs to be cropped to a square. See Fig.~\ref{fig_crop}.
\item Use the drawing tool to remove the desired location where the coin shall be placed (Fig.~\ref{fig_prompt}).
The removed background will later be filled through inpainting.
\item Provide a prompt that describes the scene and the coin.
\item DALL-E will create several images, simply select a desired image.
AI generated images are not perfect yet.
If the results are not satisfying, create additional variants and try playing around with the textual description.
No further image editing should be necessary.
DALL-E will inpaint the removed background, align the coin in a realistic perspective and add shadows that match the light source.
\item The image with the synthetic coin can now be used in a manuscript.
The article should adhere to the content policy and indicate \emph{``that the content is AI-generated in a way no user could reasonably miss or misunderstand''} \cite{DALL-E_publication_policy}.
We suggest to state in the figure caption that the size comparison is done with a fictive coin that was generated using this method.
\end{enumerate}

Current generation text-to-image generators sometimes fail to add the desired coin.
Fig.~\ref{fig_cavity_empty} shows an example where the deleted area was inpainted, but no coin was added.
It also happens that nonsensical labels are created in of object depictions (see Fig.~\ref{fig_cavity_label}).
Fig.~\ref{fig_results} shows successfully edited version of the image.
Adding coins next to the device as in Fig.~\ref{fig_cavity_coin} will likely be the desired output for this method, but sometimes surprising renderings as in Fig.~\ref{fig_cavity_hand} are generated.
There, the prompt ``add a coin that shows a sunflower on the reverse'' shows the act of fingertips actively placing a yellowish planchet next to the cavity.

\begin{figure}[ht]
\centering
\subfloat[]{\includegraphics[width=0.24\textwidth]{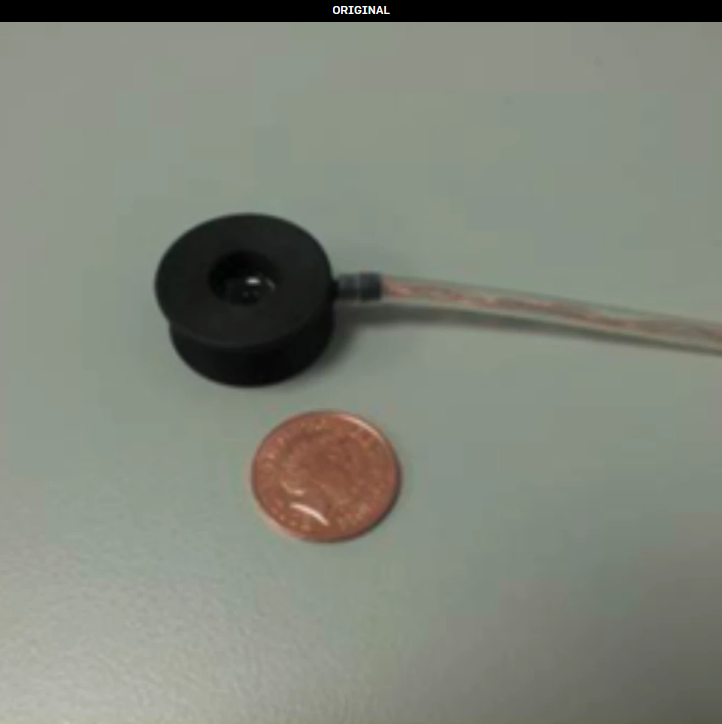}} ~
\subfloat[]{\includegraphics[width=0.24\textwidth]{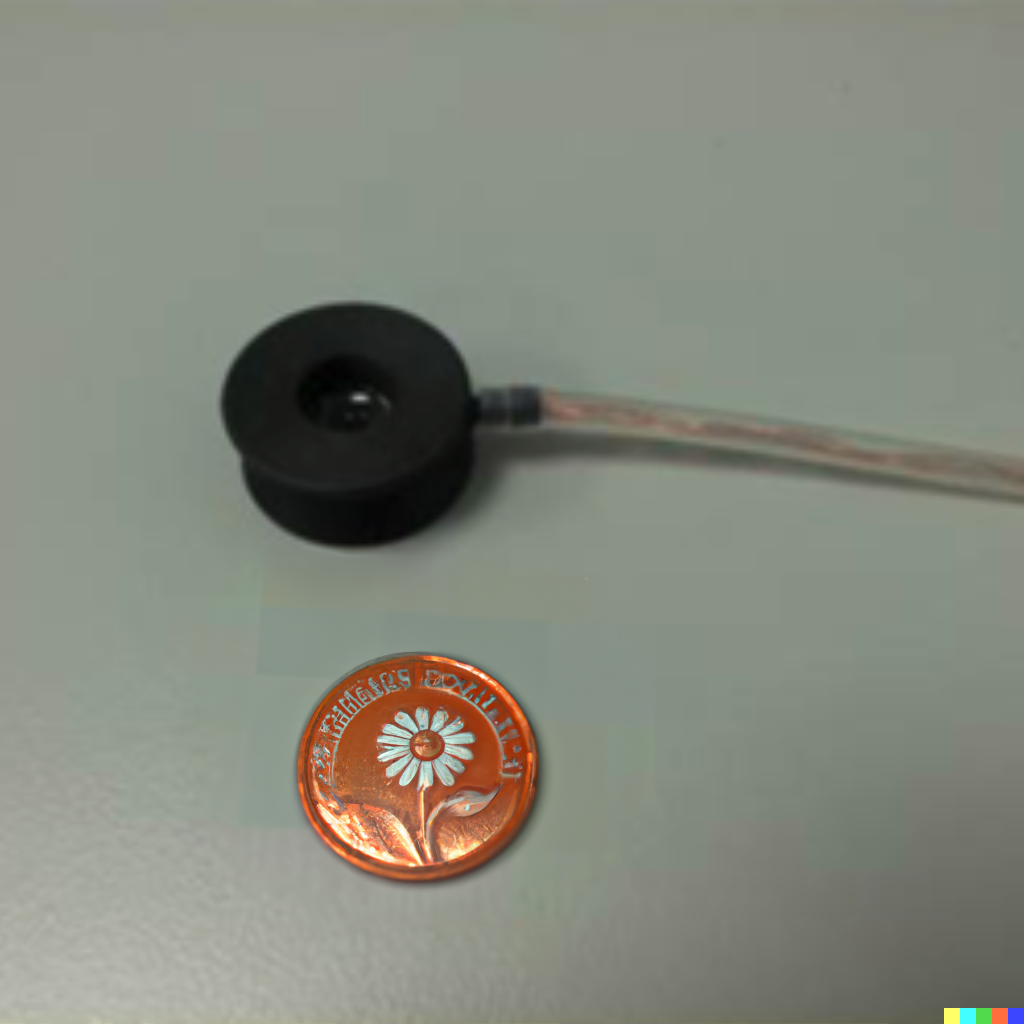}} \\
\subfloat[]{\includegraphics[width=0.24\textwidth]{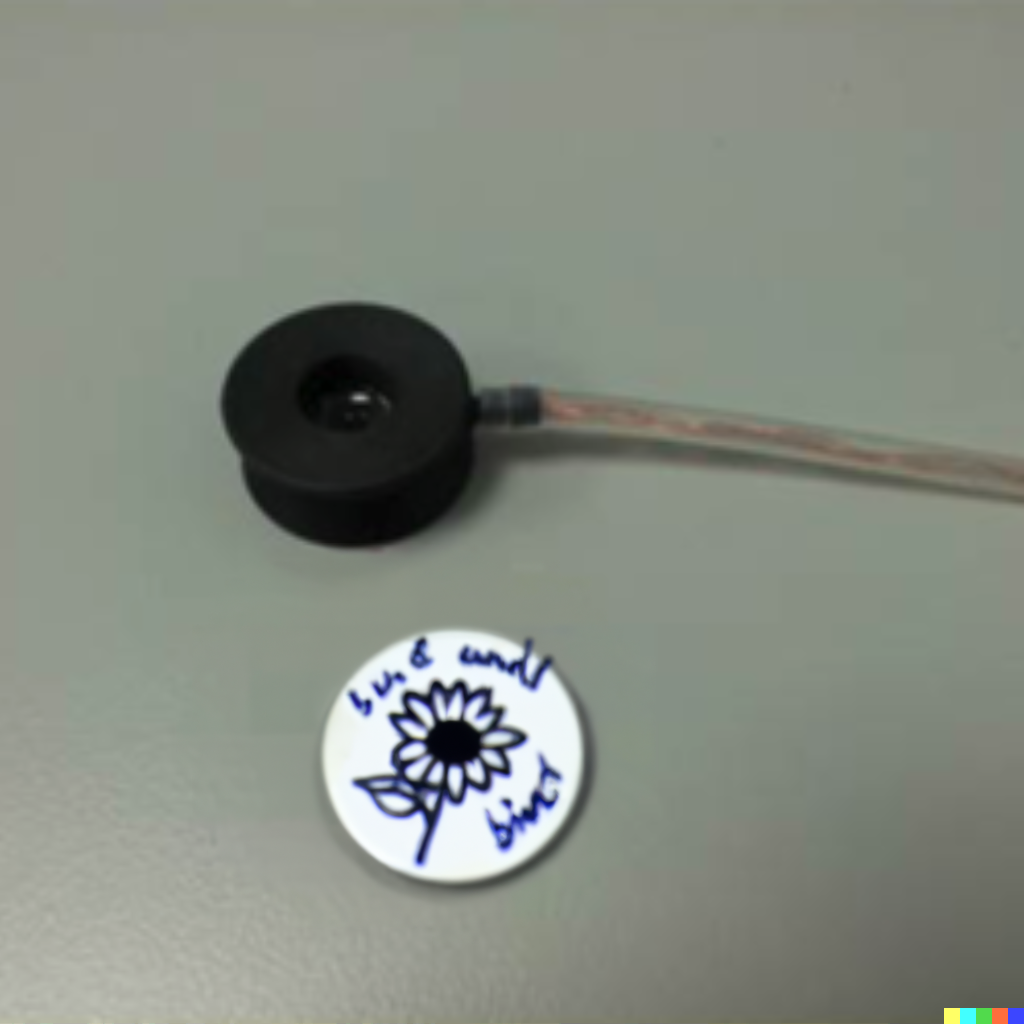}} ~
\subfloat[]{\includegraphics[width=0.24\textwidth]{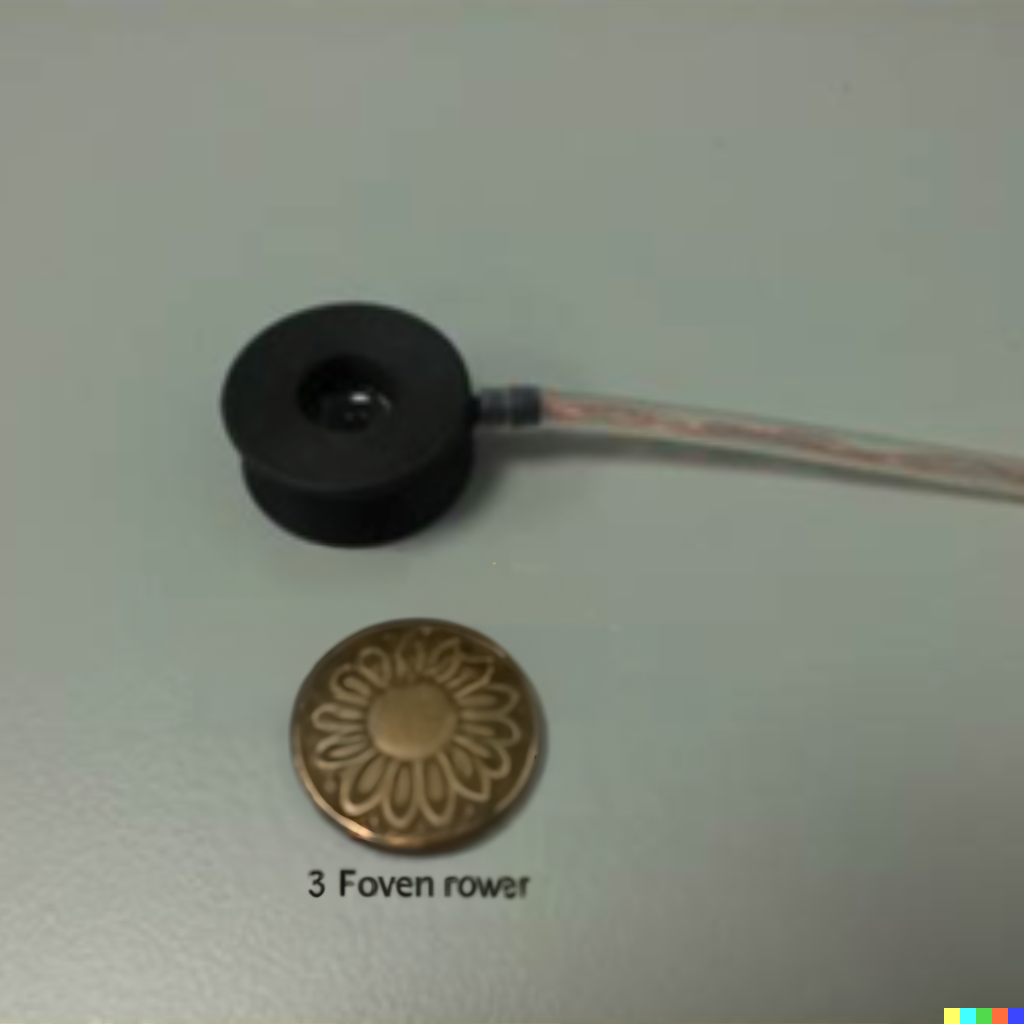}\label{fig_sunflower_d}}
\caption{Examples of a real coin exchanged with fictive AI-generated medals by DALL-E 2.
a) Original image from \cite{Lee2016}.
b) to d) Variants created by DALL-E 2 based on the uploaded original and the prompt ``add a coin that shows a sunflower on the reverse''.}
\label{fig_generated_flowers}
\end{figure}

\begin{figure}[ht]
\centering
\subfloat[]{\includegraphics[width=0.24\textwidth]{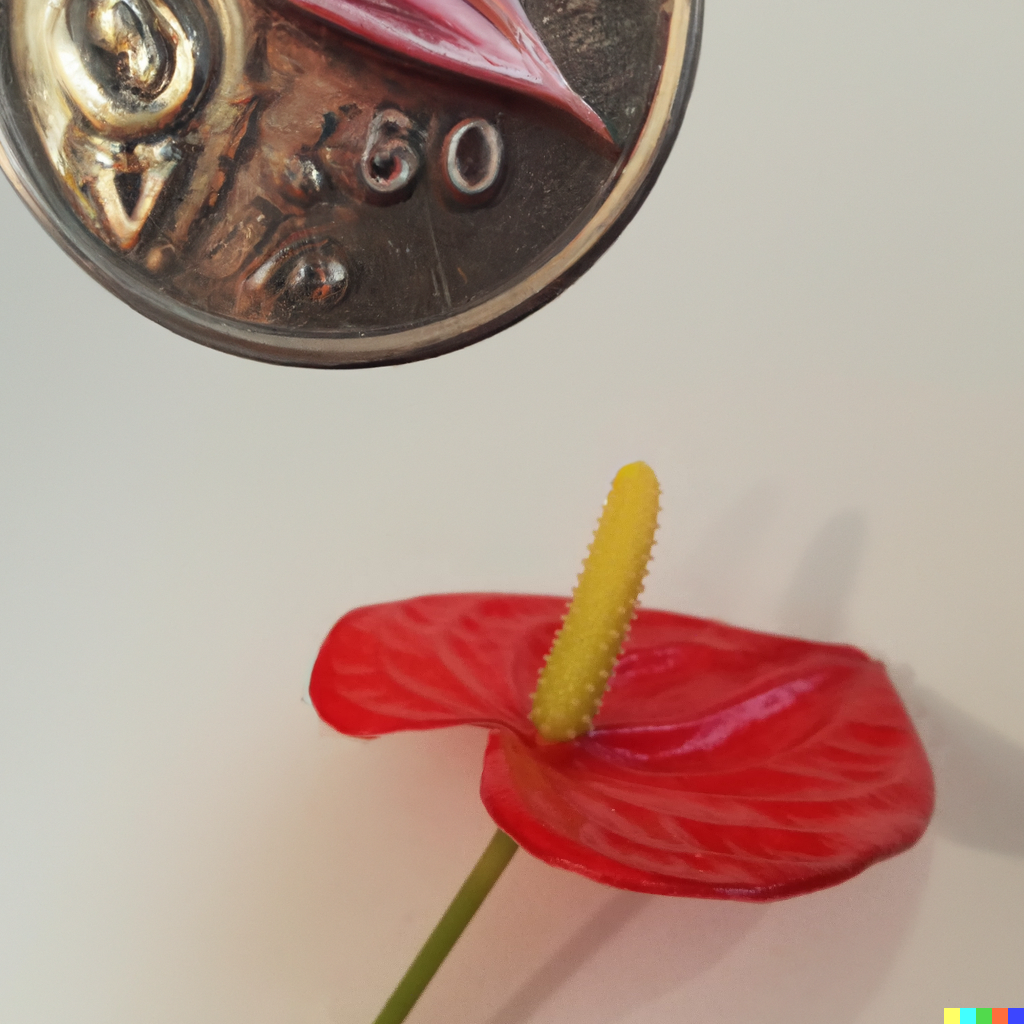}} ~
\subfloat[]{\includegraphics[width=0.24\textwidth]{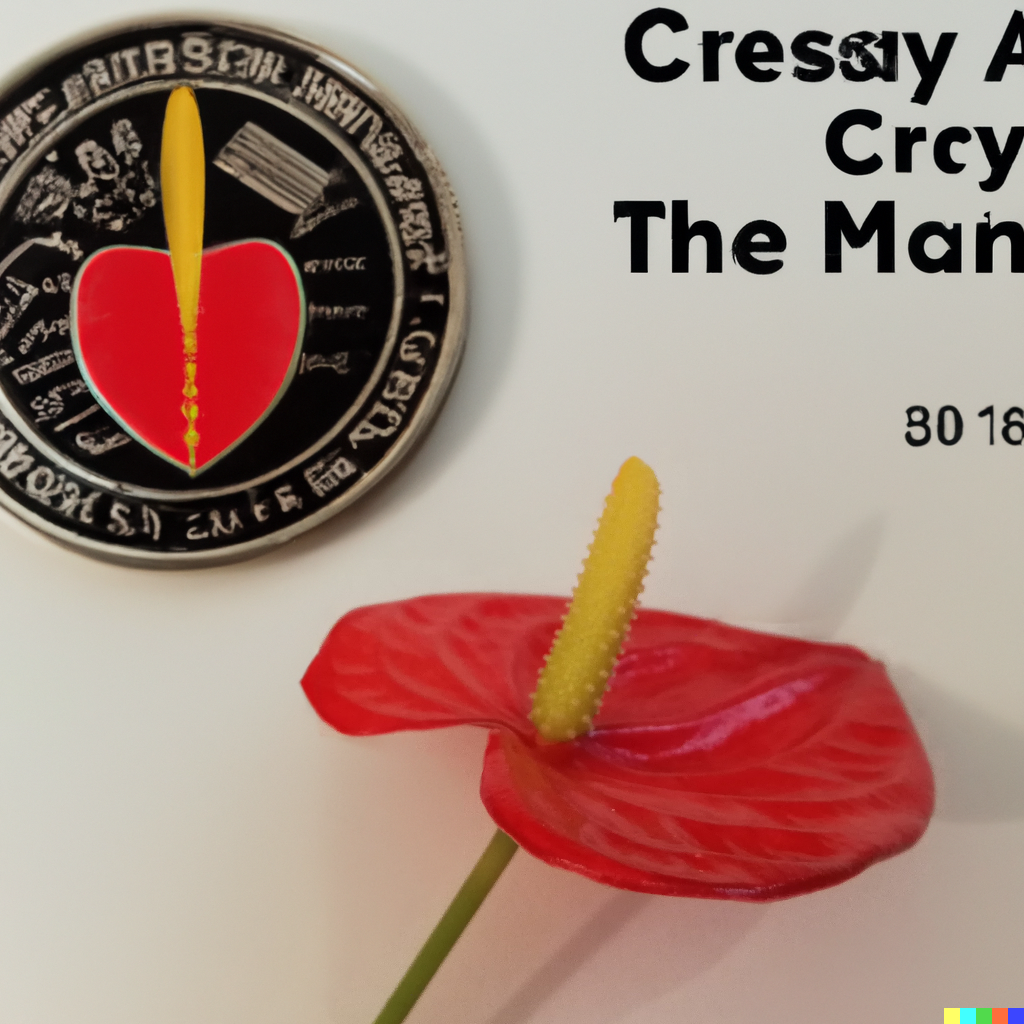}} \\
\subfloat[]{\includegraphics[width=0.24\textwidth]{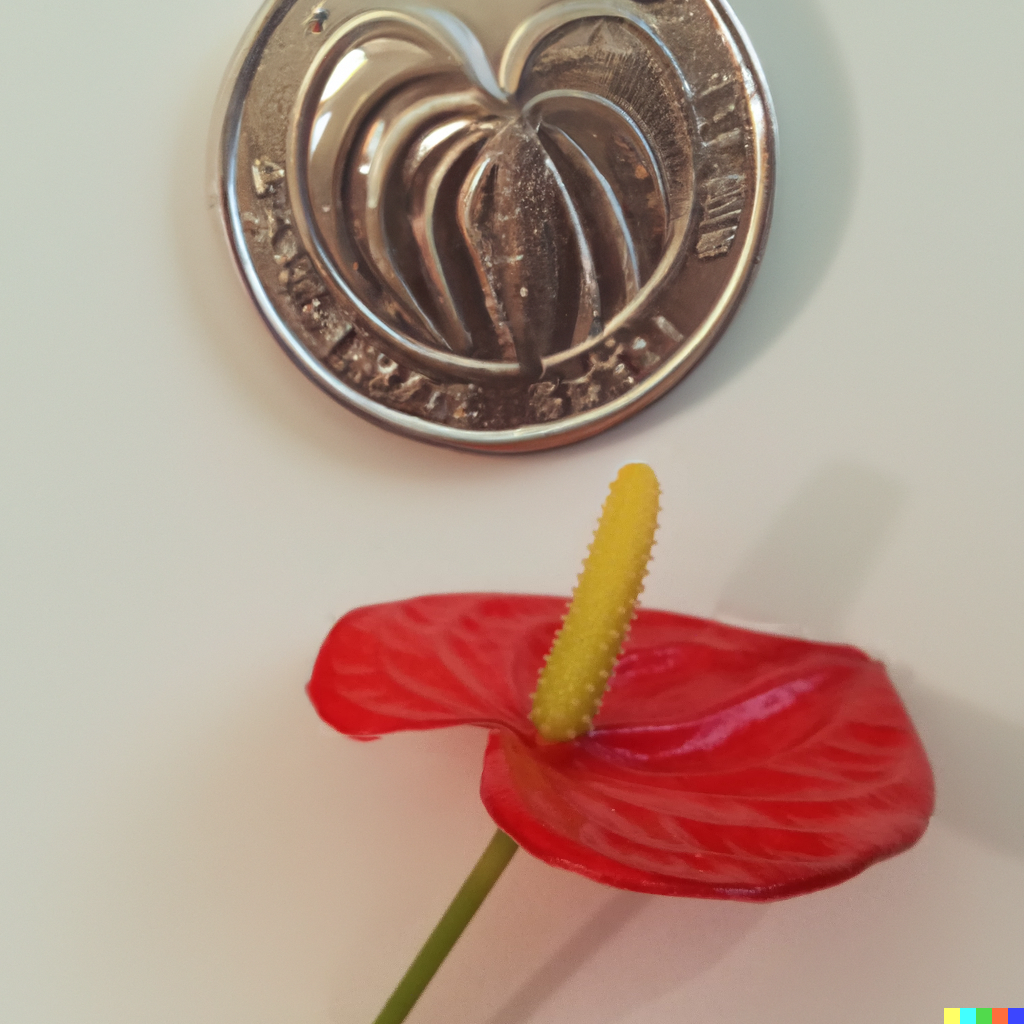}\label{fig_anthurium_realistic}} ~
\subfloat[]{\includegraphics[width=0.24\textwidth]{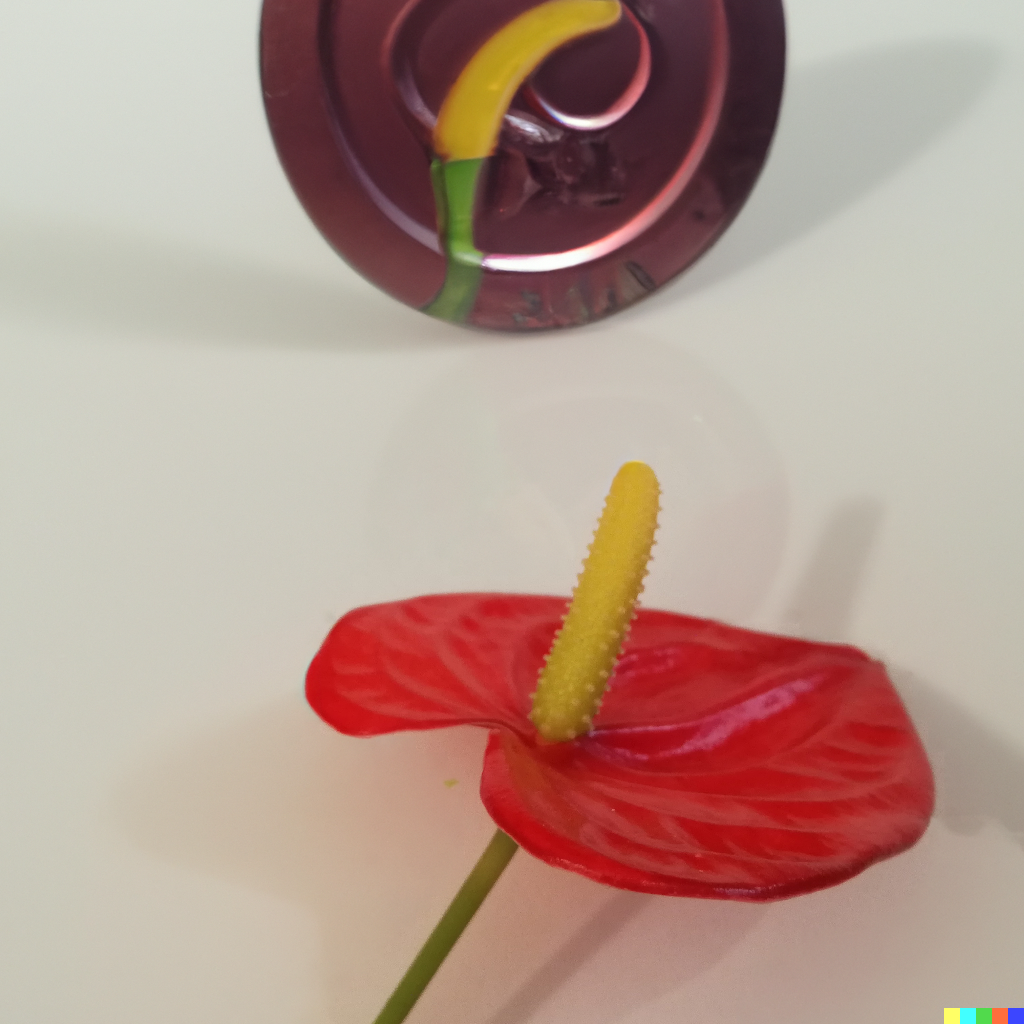}}
\caption{Selected images generated with DALL-E 2, where the face design was matched to the shown flamingo flower.
Prompts:
a) ``a euro coin in the background that shows a laceleaf'',
b) ``a coin that shows tulips on the reverse, neon'',
c) ``a coin that shows an Anthurium on the reverse, silver'',
d) ``a coin that shows an Anthurium on the reverse drawn in the style of a cartoon''.}
\label{fig_generated_laceleaf}
\end{figure}

The proposed method can enhance already existing coins in photos.
Fig.~\ref{fig_generated_flowers} shows various AI generated medals that depict sunflowers.
Using exact numismatic terminology, the synthetic images show \emph{medals} and not coins, because the fictive pieces are not legal tender.
The fictive medals don't need to have the exact size of a specific real-world coin, because the goal is to provide a rough dimension comparison and real coins also vary in size.
However, care should be taken that the medals in the generated image is indeed coin-sized to properly provide an estimate of the object's size.
The proposed technique can also match the face motif to represent the scientific field or the subjects that are being studied.
Examples are given in Fig.~\ref{fig_generated_laceleaf} where digital medals showing flamingo flowers are generated next to a photo of a real flamingo flower. 

\begin{figure}[ht]
\centering
\subfloat[]{\includegraphics[width=0.24\textwidth]{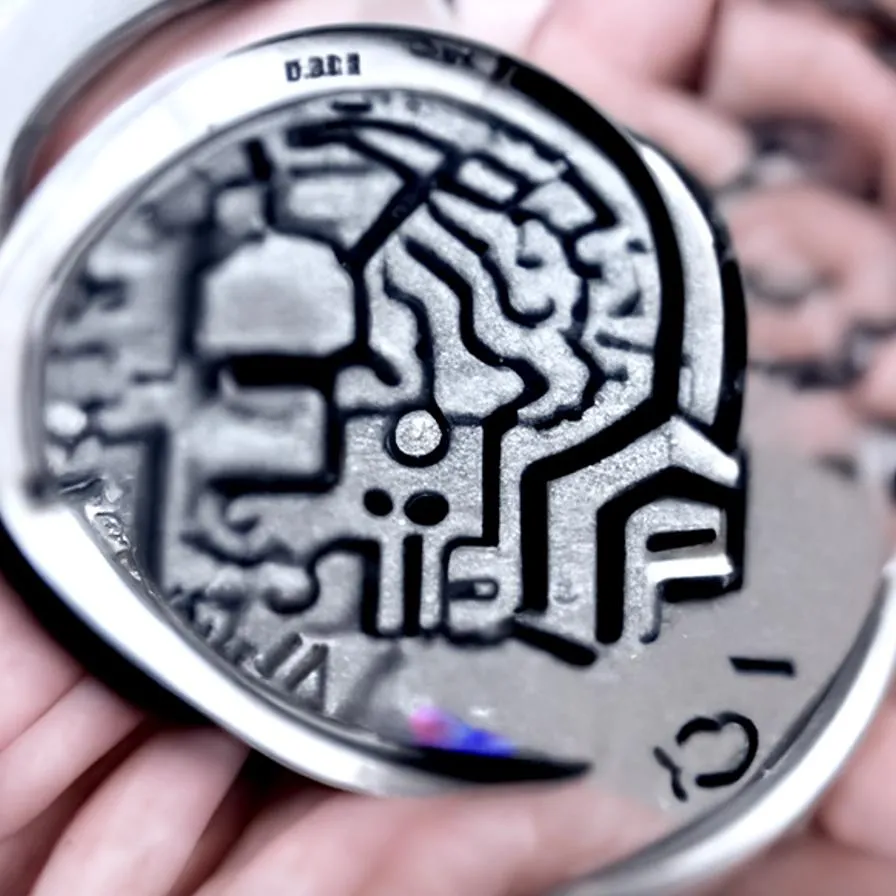}}
 ~
\subfloat[]{\includegraphics[width=0.24\textwidth]{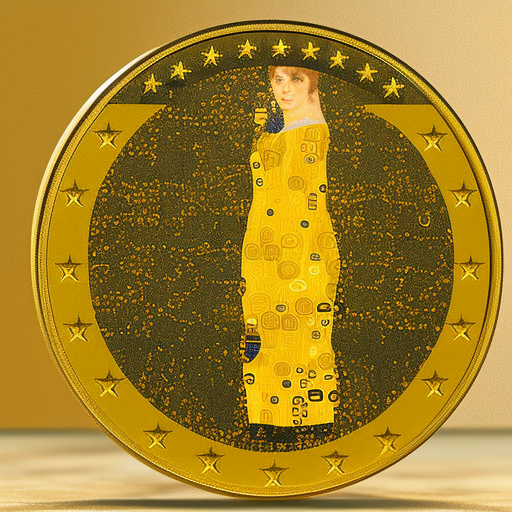}}
\\
\subfloat[]{\includegraphics[width=0.24\textwidth]{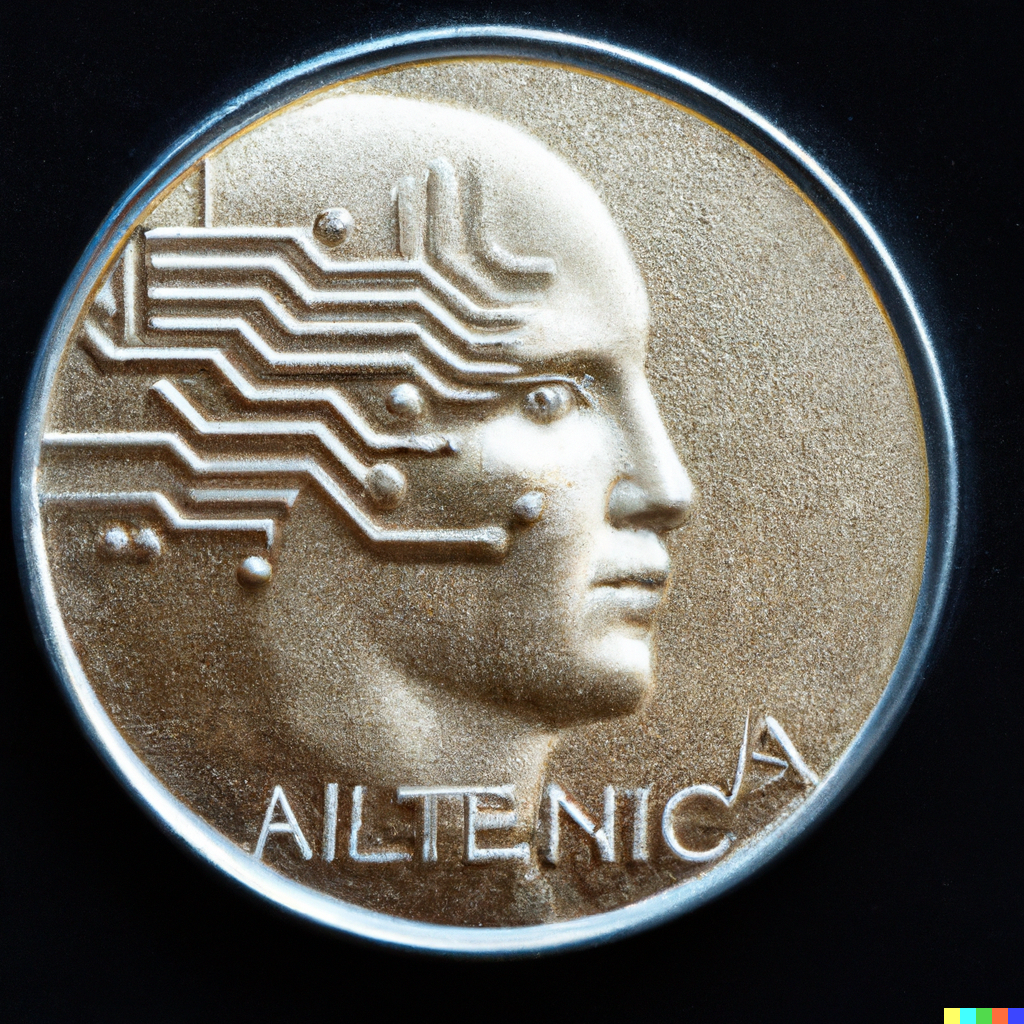}\label{fig_ai_dalle}}
 ~
\subfloat[]{\includegraphics[width=0.24\textwidth]{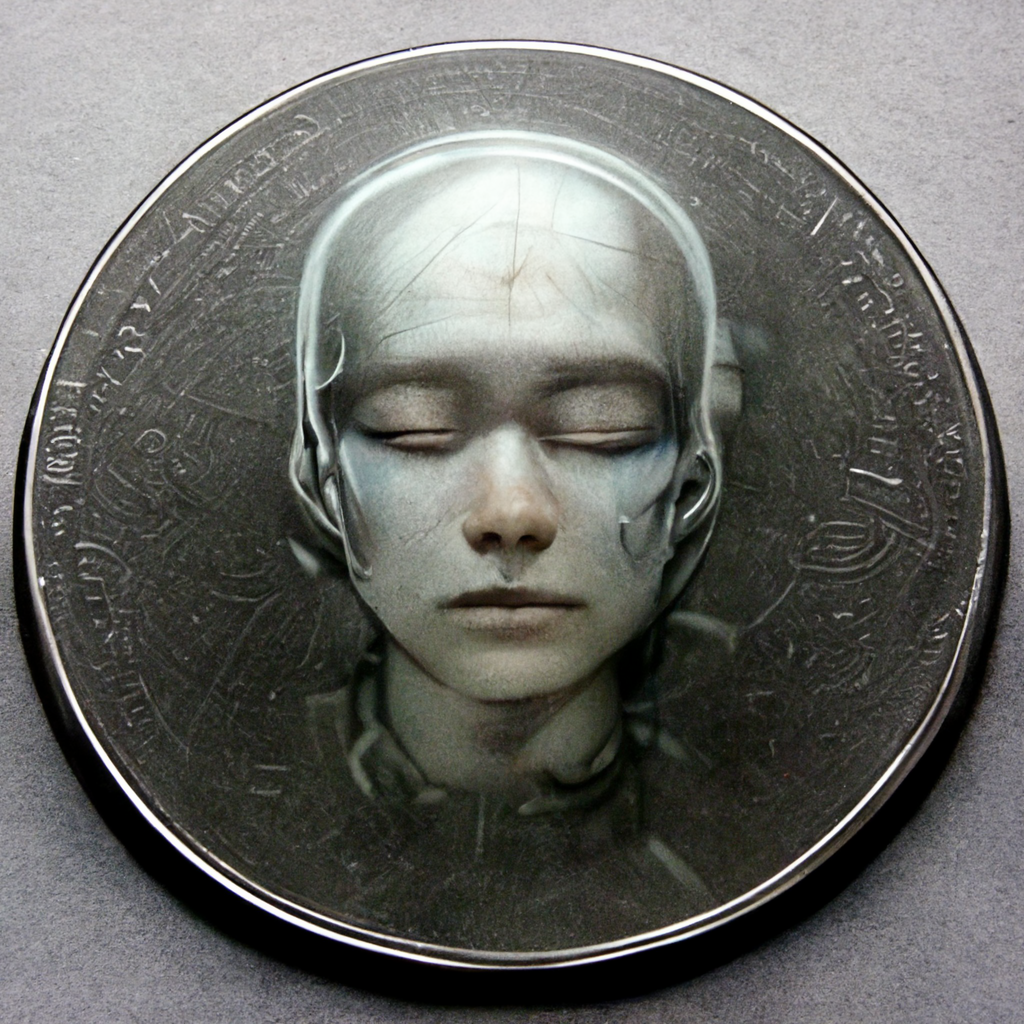}}
\caption{We have instructed different generators to synthesize coins that show artificial intelligence.
a) Night Cafe \cite{NightCafe} (VQGAN+CLIP \cite{Esser2021}) ``A coin that shows artificial intelligence''
b) Stable Diffusion: \cite{StableDiffusion} ``a hyperrealistic euro coin that shows artificial intelligence in the style of Gustav Klimt''
c) DALL-E 2: ``a coin that shows artificial intelligence on the obverse'' \cite{DALL-E}
d) Midjourney: ``A silver coin that shows artificial intelligence, lying on a table, hyper-realistic, photo'' \cite{midjourney}}

\label{fig_AI_coins}
\end{figure}

We have also asked different text-to-image generators to design coins that portrait artificial intelligence. 
The results are given in Fig.~\ref{fig_AI_coins} and vary widely from faces that are combined with electronic circuit elements to abstract geometrical forms.

\section{Conclusion}
Text-conditional image generators such as DALL-E 2 are now easy to use and require no prior knowledge of digital image processing or signal processing algorithms.
We've learned how to use these tools to add, enhance and customize coins (medals) for size reference in photos.
The method allows more creativity and personalization than circulation coins.

The numismatic capabilities of current generation text-to-image generators are limited which is likely a direct result of lacking training data in this field.
For example, obverse and reverse faces might not carry any meaning and generated images don't display numbers for monetary values or years of minting.
Texts are nonsensical throughout the synthetic images.
The images placed on coins are in some cases too complex---while colored coins are now technically feasible and production is economically viable on circulating collectors coins such as Canadian quarters, photo-realistic color images are untypical on coins.
Generative networks struggle to simultaneously fulfill a larger number of requirements.
Generating images of coins that show a specific object produce meaningful results more frequently than prompts to generate coins that lie in an area, depict an object, show a specific face and imitate a currency style.

Nevertheless, significant progress has happened for generated images of coins.
Most depictions of coins are now coherent round objects.
The perspective and shadows adopt when coins are placed in a scene.
GANs can now make sense of statements that a coin shows an object and the object is then indeed displayed on the coin and not just alongside of it.
The coin faces are already of an expected complexity and style in many generated images (see Figs.~\ref{fig_sunflower_d}, \ref{fig_anthurium_realistic} and \ref{fig_ai_dalle}).
Inpainting of surfaces is realistic and difficult to notice.




\nocite{*}


\medskip

\end{document}